# How Well Do LLMs Understand Tunisian Arabic?

**Mohamed MAHDI,** *Engineering Student*
(mohamed.mahdi@etudiant-enit.utm.tn)

*National Engineering School of Tunis, Tunis, Tunisia*

Large Language Models (LLMs) are the engines driving today's AI agents. The better these models understand human languages, the more natural and user-friendly the interaction with AI becomes, from everyday devices like computers and smartwatches to any tool that can act intelligently. Yet, the ability of industrial-scale LLMs to comprehend low-resource languages, such as Tunisian Arabic (Tunizi), is often overlooked. This neglect risks excluding millions of Tunisians from fully interacting with AI in their own language, pushing them toward French or English. Such a shift not only threatens the preservation of the Tunisian dialect but may also create challenges for literacy and influence younger generations to favor foreign languages. In this study, we introduce a novel dataset containing parallel Tunizi, standard Tunisian Arabic, and English translations, along with sentiment labels. We benchmark several popular LLMs on three tasks: transliteration, translation, and sentiment analysis. Our results reveal significant differences between models, highlighting both their strengths and limitations in understanding and processing Tunisian dialects. By quantifying these gaps, this work underscores the importance of including low-resource languages in the next generation of AI systems, ensuring technology remains accessible, inclusive, and culturally grounded.

## 1 Introduction

The future of technology will not be built on buttons or code. For the first time in human history, machines are learning to understand us through our own language. Words, expressions and dialects are becoming the new bridge between people and technology. Language models now stand at the centre of this transformation, powering intelligent agents that can plan, remember and act through conversation. This shift marks a turning point: our native language is no longer just a tool for human interaction, it is becoming the interface of the digital world.

In this new era, how well a machine understands language determines how much it can serve its users. But what happens when that language is not one of the global standards, not English, French, or Modern Standard Arabic, but a vibrant, living dialect like Tunisian Arabic? Spoken by over twelve million people across Tunisia, this dialect carries centuries of culture, history and emotion. Yet it remains one of the least represented languages in modern AI systems. Its diversity of accents, the lack of a unified written form, and the coexistence of two scripts, Latin ("Tunizi") and Arabic, make it a formidable challenge for today's large-language models. Words can be spelled in dozens of ways, mixed with Amazigh, French, Turkish, and English roots, reflecting Tunisia's layered and complex past.

This challenge is not just technical. It is cultural and social. If modern AI systems cannot understand our dialect, Tunisians may be forced to switch to foreign languages simply to interact with their own devices and digital agents. That risk touches literacy, accessibility and identity: younger generations might become more comfortable speaking to machines in English or French than in their mother tongue, gradually distancing themselves from native expression. Ensuring that future AI can speak and understand Tunisian Arabic is therefore not only a matter of innovation, it is a matter of inclusion, preservation and technological equity.

Although there has been growing interest in Arabic NLP, few efforts focus specifically on the Tunisian dialect. For example, the DIALEX benchmark covers several Arabic dialects, including Tunisian[1]. The TARC corpus collects Tunisian Arabizi (Latin-script dialect) texts for computational and linguistic analyses[2]. More recently, TunBERT was introduced as the first monolingual Transformer-based language model trained for Tunisian dialect and evaluated on sentiment analysis, dialect identification and reading-comprehension tasks[3].

In this paper, we explore how well today's major language models can understand the Tunisian dialect. We introduce a manually crafted dataset of one hundred examples: each item includes a sentence in Tunizi, its equivalent in Arabic script, its English translation, and a sentiment label (positive, negative or neutral). Using this dataset, we benchmark several models across three tasks: transliteration (Latin → Arabic), translation (Tunisian Arabic/Tunizi → English) and sentiment classification. For translation and sentiment tasks, we employ standard metrics: BLEU[5], METEOR[6], and BERTScore[7], to measure model performance in both surface and semantic space. Our findings reveal that, while some understanding exists, models remain far behind their performance with Standard Arabic, English or better-resourced dialects.

Our goal is not only to report these results, but to draw attention to the gap, motivate larger datasets and evaluation efforts, and highlight that linguistic inclusivity matters when building the next generation of AI systems. This study serves as an initial exploration and proof-of-concept benchmark for evaluating LLMs on Tunisian Arabic. Future work will expand the dataset to support statistically robust comparisons. The rest of this paper is organised as follows: Section 2 presents the dataset, its collection and annotation, and details the methodology and experimental setup used to evaluate language models. Section 3 discusses the results. Section 4 concludes with implications and future directions for Tunisian dialect understanding in large-language models.

All datasets, task-specific prompts, and evaluation scripts used in this study are available in our public repository[1], along with the raw model outputs obtained for each experiment.

## 2 Dataset and Methodology

### 2.1 Dataset Collection

The dataset was collected from social media comments, including platforms such as Facebook, X, YouTube, Instagram, and discussion forums. These platforms were chosen because they reflect the diversity of Tunisian populations and capture everyday language use. By collecting from multiple sources, we aimed to cover a broad range of topics, expressions, and writing styles that Tunisian speakers naturally use online. Unlike existing resources such as TArC[2], CTAB[4], or TunBERT[3], which may already be included in the pretraining data of industrial LLMs, our dataset was manually curated from recent, non-indexed social media content. This ensures that evaluation results reflect genuine model generalization rather than memorization of known Tunisian data.

Comments were selected according to several criteria. First, we included comments from different regions of Tunisia, including the north, south, coastal, and inland areas. Second, we sought a variety of sociocultural topics, such as daily life, politics, humor, family, slang, and traditional culture. Third, comments with different registers were included: informal chat, mixed-language expressions (Arabic, French, English), and code-switching.

Ethical considerations were central in the collection process. All data was anonymized, and any personally identifiable information, such as usernames or links, was removed to ensure privacy.

### 2.2 Data Preprocessing

The collected comments were manually filtered to include only those written in a readable format, preserving their original handwritten style. Emojis and non-textual content were removed, and only comments written entirely in *Tunizi* (Latin-script Tunisian Arabic) were retained. This ensured that the dataset reflects authentic language use without altering its natural variation or informal characteristics.

### 2.3 Annotation

Each comment in the dataset was annotated for three tasks: transliteration (from Tunizi to Arabic script), translation (Tunisian Arabic to English), and sentiment classification (Positive, Negative, Neutral). The annotation process was carried out manually by a single annotator (the author), ensuring consistency and high quality.

Table 1 shows three illustrative examples from the dataset:

| Tunizi | Arabic | English | Sentiment |
|---|---|---|---|
| fesh takra? | فاش تقرأ؟ | What are you studying? | NEUTRAL |
| Wallah chay yedaoekh | والله شيئ يدوخ | I swear, this is mind blowing | NEGATIVE |
| el padel paddeli hyeti | البادل، بدلي حياتي | Padel changed my life | POSITIVE |

Table 1: Sample annotations from the dataset, showing Tunizi, Arabic script, English translation, and sentiment labels.

---

[1] https://github.com/Mahdyy02/llm-tunisian-language

These examples demonstrate the diversity of expression, script, and sentiment present in the dataset, while also highlighting the challenges of transliteration and translation for modern language models.

### 2.4 Dataset Statistics

The final dataset contains 100 samples. Among these, 37% are labeled as *Positive*, 35% as *Negative*, and 27% as *Neutral*. Figure 1 illustrates the sentiment distribution, providing a clear visual overview of the dataset's balance.

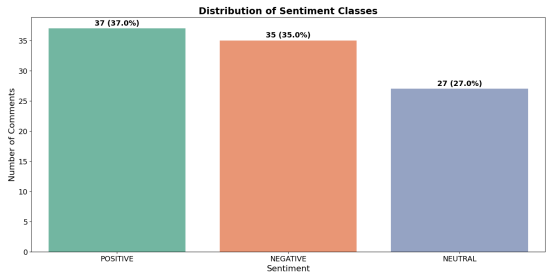

Fig. 1: Histogram showing the distribution of sentiment classes in the dataset.

Tunisian Arabic presents several challenges that make it a compelling target for LLM evaluation. Written content is scarce and highly inconsistent, reflecting regional dialects, personal writing styles, and frequent code-switching with French and English. This linguistic diversity enriches the dataset but simultaneously introduces complexity for language models. Moreover, the lack of a standardized written form amplifies these challenges, making it difficult for models to generalize across different spellings, slang, and syntactic patterns.

### 2.5 Methodology Overview

To rigorously evaluate the ability of modern language models to understand Tunisian Arabic, we designed a three-task benchmark covering transliteration, translation, and sentiment classification. Each task targets a specific aspect of language understanding, ensuring a comprehensive assessment.

**Transliteration.** The first task measures how well models can convert Tunizi text, written in Latin characters, into standard Arabic script. This is challenging because there is no fixed spelling system for Tunisian Arabic, and words often vary across regions. To quantify performance, we computed three complementary metrics: Character Error Rate (CER), Levenshtein distance, and Longest Common Subsequence (LCS) similarity. CER captures the fraction of characters that differ between the predicted and ground truth sequences. Levenshtein distance counts the minimal number of edits required to match the sequences, while LCS evaluates the longest sequence of matching characters in order, providing a sense of overall structural similarity. These metrics together provide a nuanced view of transliteration quality, from character-level accuracy to sequence coherence.

**Translation.** The second task evaluates how accurately models translate Tunisian Arabic into English. We employed three widely used metrics in machine translation research: BLEU, METEOR, and BERTScore. BLEU measures n-gram overlap between the model output and reference translations, focusing on exact word matches. METEOR incorporates stemming and synonym matching, offering a more flexible assessment of meaning preservation. BERTScore uses deep contextual embeddings to measure semantic similarity, capturing meaning even when words differ. We tokenize both reference and predicted sentences to ensure consistent evaluation across all models.

**Sentiment Classification.** The third task assesses the ability of models to correctly classify the sentiment of each comment as *Positive*, *Negative*, or *Neutral*. We calculated standard classification metrics, including accuracy, weighted and macro-averaged precision, recall, and F1-score. Additionally, we report Cohen's Kappa and Matthews Correlation Coefficient to measure agreement beyond chance, especially for multi-class prediction. Per-class metrics and confusion matrices were computed to analyze model performance in detail for each sentiment category.

All evaluations were performed using our manually curated dataset, ensuring high-quality ground truth for both transliteration, translation, and sentiment labels. While the dataset is modest in size, its careful annotation allows for meaningful benchmarking of LLMs, revealing both their current capabilities and the specific challenges posed by Tunisian Arabic. The code and evaluation pipeline are fully reproducible, providing a foundation for future studies and dataset expansion.

## 3 Results and Analysis

In this section, we present the performance of several large language models (LLMs) on the three evaluation tasks: Tunisian-Arabic transliteration (Task 1), translation to English (Task 2), and sentiment classification (Task 3). All experiments were conducted using the latest versions of the models as of November 9, 2025. We focus on highlighting the best-performing models for each task while providing a comparative overview of all evaluated LLMs. Given the dataset's size, results are indicative rather than statistically conclusive; scores are reported on the full set for exploratory benchmarking.

## 3.1 Task 1: Tunisian-Arabic Transliteration

Table 2 summarizes the transliteration performance of the tested LLMs. We report the Character Error Rate (CER), Levenshtein distance, and Longest Common Subsequence similarity (LCS). Lower CER and Levenshtein values indicate better performance, while higher LCS indicates better alignment with the ground truth.

Among all tested systems, **Gemini 2.5 Flash** achieves the best overall performance with a CER of **0.15**, LEV of **2.89**, and LCS of **0.88**. These results confirm Gemini's strong capability to reproduce Tunisian utterances accurately in Arabic script, capturing fine-grained phonetic details while maintaining orthographic consistency. This can be attributed to its extensive multilingual pre-training and recent fine-tuning updates (as of November 2025), which enhance cross-script generalization.

**Claude Sonnet 4.5** and **DeepSeek-R1** also exhibit competitive results, with CERs of **0.19** and **0.20**, respectively. Their relatively low Levenshtein distances (**3.78** and **4.02**) and high LCS values (**0.86** and **0.85**) suggest that both models handle phoneme-to-grapheme mapping effectively, though they occasionally introduce minor inconsistencies in morphological endings or diacritics. These deviations likely arise from subtle dialectal variations not well represented in their multilingual corpora.

The mid-tier group includes **Grok 3** (CER **0.23**, LEV **4.43**, LCS **0.83**) and **GPT-4o Mini** (CER **0.26**, LEV **5.19**, LCS **0.80**). Although both models exhibit reasonable accuracy, they tend to normalize colloquial Tunisian tokens into Modern Standard Arabic or English transliterations, which increases edit distance despite semantically correct outputs.

In contrast, **Mistral 8×22B** and **Qwen 3** perform poorly, with CERs of **1.29** and **3.16**, and notably high Levenshtein distances (**23.64** and **53.81**). Their very low LCS scores (**0.26** and **0.24**) indicate that these open-source models struggle substantially with informal North-African Arabic orthographies. This is likely due to the lack of dialectal or phonetic Arabic data in their training sets and limited exposure to code-switched content.

Overall, these results highlight the importance of broad multilingual coverage and dialect-aware fine-tuning in cross-script generation tasks. Gemini's superior performance demonstrates how continuous large-scale updates and diversified data sources can significantly enhance the ability of LLMs to model under-represented dialects such as Tunisian Arabic.

## 3.2 Task 2: Translation to English

For translation evaluation, Table 3 presents BLEU, METEOR, and BERTScore F1 metrics. Higher values indicate better translation quality.

**Gemini 2.5 Flash** once again achieves the best overall results, with a METEOR score of **0.45** and a BERTScore F1 of **0.91**, confirming its superior ability to preserve both lexical and semantic alignment with human translations. Although its BLEU score (**0.11**) is comparable to that of Claude, the higher METEOR and BERTScore values indicate that Gemini captures meaning beyond surface-level word overlap, an essential quality for dialect translation tasks.

**Claude Sonnet 4.5** and **GPT-4o Mini** follow closely behind, achieving METEOR scores of **0.37** and **0.37** and BERTScore F1 values of **0.90** and **0.90**, respectively. These results suggest both models demonstrate a reasonable understanding of Tunisian semantics and idiomatic usage, but occasionally normalize dialectal expressions into Modern Standard Arabic or omit context-specific nuances.

**DeepSeek-R1** performs slightly lower across all metrics (BLEU **0.09**, METEOR **0.35** and BERTScore F1 **0.90**), indicating that despite strong reasoning capabilities, its translation pipeline struggles with the informal and phonetically inconsistent nature of Tunisian dialect.

Conversely, **Grok 3**, **Mistral 8×22B**, and **Qwen 3** display notably weak results, with METEOR scores around **0.09–0.10** and BERTScore F1 below **0.86**. Their low BLEU values (**0.02**) reveal frequent mistranslations and semantic drift, often producing unrelated or overly literal outputs. This reflects the absence of North-African dialect data in their training corpora and limited adaptability to hybrid Arabic–French lexical forms commonly found in Tunisian online text.

Overall, these results emphasize that while modern proprietary LLMs (Gemini, Claude, GPT) demonstrate solid generalization to low-resource dialects, open-source models still require targeted fine-tuning with dialect-specific parallel data to reach competitive translation fidelity.

## 3.3 Task 3: Sentiment Classification

Table 4 presents the sentiment classification performance of various LLMs on our Tunisian dataset, including the versions used as of November 2025.

Overall, **GPT-4o Mini** achieves the best overall performance, with the highest **Accuracy (0.60)**, **Weighted F1 (0.60)**, **Macro F1 (0.45)**, and **F1 for Positive sentiment (0.69)**. This suggests GPT-4o Mini is particularly strong at correctly classifying

| LLM | Version (Nov 2025) | CER | Levenshtein | LCS |
|---|---|---|---|---|
| Claude | Claude Sonnet 4.5 | 0.19 | 3.78 | 0.86 |
| DeepSeek | DeepSeek-R1 | 0.21 | 4.02 | 0.85 |
| Gemini | Gemini 2.5 Flash | **0.15** | **2.89** | **0.88** |
| GPT | GPT-4o Mini | 0.26 | 5.19 | 0.80 |
| Grok | Grok 3 | 0.23 | 4.43 | 0.83 |
| Mistral | Mixtral 8×22B | 1.29 | 23.64 | 0.26 |
| Qwen | Qwen 3 | 3.16 | 53.81 | 0.24 |

Table 2: Transliteration performance of LLMs on Tunisian-Arabic text, including the versions used (as of November 2025). Best values are highlighted in bold.

| LLM | Version (Nov 2025) | BLEU | METEOR | BERTScore F1 |
|---|---|---|---|---|
| Claude | Claude Sonnet 4.5 | **0.12** | 0.37 | 0.90 |
| DeepSeek | DeepSeek-R1 | 0.09 | 0.35 | 0.90 |
| Gemini | Gemini 2.5 Flash | 0.11 | **0.45** | **0.91** |
| GPT | GPT-4o Mini | 0.10 | 0.37 | 0.90 |
| Grok | Grok 3 | 0.02 | 0.10 | 0.85 |
| Mistral | Mixtral 8×22B | 0.02 | 0.10 | 0.85 |
| Qwen | Qwen 3 | 0.02 | 0.09 | 0.85 |

Table 3: Translation performance of LLMs from Tunisian-Arabic to English, including versions used (as of November 2025). Best values are highlighted in bold.

| LLM | Version (Nov 2025) | Accuracy | F1 Weighted | F1 Macro | F1 Positive | F1 Negative | F1 Neutral | Cohen Kappa | Mcc | Precision Positive | Precision Negative | Precision Neutral |
|---|---|---|---|---|---|---|---|---|---|---|---|---|
| Claude | Claude Sonnet 4.5 | 0.58 | 0.58 | 0.44 | 0.6 | 0.52 | **0.64** | 0.37 | 0.37 | 0.56 | 0.55 | 0.65 |
| Deepseek | DeepSeek-R1 | 0.5 | 0.5 | 0.38 | 0.52 | 0.49 | 0.51 | 0.26 | 0.26 | 0.59 | 0.49 | 0.44 |
| Gemini | Gemini 2.5 Flash | 0.46 | 0.46 | 0.35 | 0.38 | 0.52 | 0.49 | 0.2 | 0.21 | 0.42 | **0.62** | 0.49 |
| Gpt | GPT-4o Mini | **0.6** | **0.6** | **0.45** | **0.69** | 0.54 | 0.56 | **0.41** | **0.41** | 0.71 | 0.61 | 0.49 |
| Grok | Grok 3 | 0.51 | 0.47 | 0.36 | 0.35 | **0.58** | 0.51 | 0.26 | 0.32 | **0.89** | 0.44 | 0.6 |
| Mistral | Mixtral 8×22B | 0.41 | 0.39 | 0.29 | 0.49 | 0.31 | 0.35 | 0.1 | 0.11 | 0.5 | 0.5 | 0.38 |
| Qwen | Qwen 3 | 0.44 | 0.37 | 0.27 | 0.29 | 0.55 | 0.26 | 0.14 | 0.22 | 0.64 | 0.55 | **1.0** |

Table 4: Sentiment classification performance of LLMs on the Tunisian dataset, including versions used (as of November 2025). Maximum values per row highlighted in bold.

the dominant Positive class while maintaining balanced performance across the other classes.

**Claude Sonnet 4.5** also demonstrates competitive performance, with high F1 for Neutral sentiment (**0.64**) and balanced metrics overall, indicating its strength in handling the more nuanced Neutral class. **Grok 3**, while not leading in overall accuracy, achieves the highest F1 for Negative sentiment (**0.59**) and very high precision for Positive sentiment (**0.89**), highlighting that some models specialize in specific classes even if their overall score is lower.

Other models, such as **DeepSeek-R1**, **Gemini 2.5 Flash**, **Mistral**, and **Qwen 3**, show lower overall performance, with particular difficulty distinguishing between Positive and Neutral classes. The lower Cohen's Kappa and MCC scores for these models reflect less agreement with the ground truth labels and potential biases toward certain classes.

These results illustrate several key points: (i) model architecture and pretraining data significantly impact the ability to handle low-resource dialects like Tunisian Arabic, (ii) even strong models may show class-specific strengths and weaknesses, and (iii) the latest model versions at the time of evaluation matter for reproducibility and future benchmarking.

### 3.4 Discussion

Across all three tasks, Gemini and GPT consistently emerge as the top-performing models, while Qwen and Mistral lag considerably. This pattern underscores the substantial variability in LLM performance when applied to low-resource dialects such as Tunisian Arabic. Even the leading models achieve lower scores compared to well-resourced languages, highlighting the pressing need for expanded datasets, dedicated benchmarks, and evaluation frameworks tailored to regional dialects.

Interestingly, Gemini achieves near-top performance across almost all tasks. While the exact reasons behind this dominance remain multifaceted and partly speculative, several factors may contribute. One possibility is the sheer volume and diversity of data accessible to the developers, including potentially more extensive multilingual corpora or web-based resources. Another consideration could be the presence of engineers and researchers with regional expertise in Arabic, which might influence model pretraining, fine-tuning, or evaluation strategies. Additionally, Gemini's integra-tion with broader Google research infrastructure may provide advantages in optimization, model iteration, and continuous learning from diverse sources.

It is important, however, to resist oversimplifying these results. Performance cannot be attributed solely to data quantity or geographic representation; architecture, pretraining objectives, fine-tuning protocols, and even evaluation alignment all play significant roles. By presenting these results, we aim not only to document the current state of LLM performance on Tunisian Arabic but also to invite readers to reflect on the interplay of data, design, and human expertise that drives model behavior. Future work could explore these factors in a controlled setting, potentially shedding light on the relative impact of data, model architecture, and human knowledge on dialectal performance.

Finally, we emphasize that all evaluations were conducted on the latest model versions available as of November 9, 2025. Given the rapid pace of LLM development, performance is likely to evolve, reinforcing the value of establishing a reproducible baseline at a specific point in time.

## 4 Conclusion

In this study, we evaluated the performance of several state-of-the-art large language models on the Tunisian Arabic dialect across three key tasks: transliteration, translation, and sentiment classification. Rather than serving as a definitive benchmark, this work represents a pilot-scale initiative to seed a national collaborative effort for Tunisian Arabic datasets. Our results show that while models like Gemini and GPT perform relatively well, overall performance remains lower than what is typically observed for well-resourced languages. Models such as Qwen and Mistral lag significantly, highlighting the persistent challenges posed by low-resource dialects.

These findings emphasize that Tunisian Arabic, despite its richness and cultural significance, remains underrepresented in the datasets that drive modern AI. If we aim for a future where LLMs are truly universal, capable of understanding and supporting every language and dialect, then national efforts to create large, high-quality datasets are essential. By investing in dedicated projects and collaborations, Tunisian researchers can ensure that our language is included in the next generation of AI systems.

We hope this work inspires the community to prioritize linguistic inclusivity, to develop open-access datasets, and to engage with LLM providers for fine-tuning models that understand the Tunisian dialect. LLMs are poised to be present in almost every digital interaction of the future; making sure Tunisian Arabic has a strong and visible presence is not just a technical challenge, it is a cultural imperative.

Together, through careful data collection, evaluation, and collaboration, we can help shape an AI ecosystem that respects and reflects the linguistic diversity of Tunisia, ensuring that our language is represented, valued, and usable in the technologies of tomorrow.